\documentclass[conference]{IEEEtran}
\IEEEoverridecommandlockouts
\usepackage{cite}
\usepackage{amsmath,amssymb,amsfonts}
\usepackage{algorithmic}
\usepackage{graphicx}
\usepackage{textcomp}
\usepackage{xcolor}
\def\BibTeX{{\rm B\kern-.05em{\sc i\kern-.025em b}\kern-.08em
    T\kern-.1667em\lower.7ex\hbox{E}\kern-.125emX}}

\usepackage{fancyhdr,graphicx,amsmath,amssymb}
\usepackage[ruled,vlined]{algorithm2e}
\include{pythonlisting}
\usepackage[margin=1in]{geometry}
\usepackage[english]{babel}
\usepackage{subcaption}
\usepackage{comment}

\newcommand{\M}{\mathcal{M}}
\newcommand{\T}{\mathcal{T}}
\usepackage[skins]{tcolorbox}

\newtheorem{example}{Example}

\newcommand{\lu}[1]{{\footnotesize\color{magenta}[{\bf Lu:} \textsf{#1}]}}

\newcommand{\startpara}[1]{{\vskip1pt\noindent{\bf #1.}}}

\begin{document}

\title{Towards Transparent Robotic Planning \\ via Contrastive Explanations
\thanks{}
}

\author{\IEEEauthorblockN{Shenghui Chen$^*$, Kayla Boggess$^*$, and Lu Feng
\thanks{$^*$Equal Contribution}  
\IEEEauthorblockA{Department of Computer Science,
University of Virginia\\
Charlottesville, USA\\
Emails: \{sc9by, kjb5we, lf9u\}@virginia.edu}}}

\maketitle

\begin{abstract}

Providing explanations of chosen robotic actions can help to increase the transparency of robotic planning and improve users' trust.
Social sciences suggest that the best explanations are contrastive, explaining not just why one action is taken, but why one action is taken instead of another. 
We formalize the notion of \emph{contrastive explanations} for robotic planning policies based on Markov decision processes, drawing on insights from the social sciences. 
We present methods for the automated generation of contrastive explanations with three key factors: selectiveness, constrictiveness and responsibility.
The results of a user study with 100 participants on the Amazon Mechanical Turk platform 
show that our generated contrastive explanations can help to increase users' understanding and trust of robotic planning policies, while reducing users' cognitive burden. 


\end{abstract}


\section{Introduction}\label{intro}

In recent years, there has been a significant amount of work done in the field of Explainable AI~\cite{arrieta2020explainable,Gunningeaay7120,adadi_berrada_2018}, to increase the transparency of AI decision-making systems and improve users’ trust.
Because of traditional ``black-box'' approaches, lay-users have little understanding of how a decision is made or why an action occurs, often leading to misunderstanding and mistrust of the system, which can further lead to problems caused by system misuse~\cite{goebel2018explainable}. 
The vast majority of work in Explainable AI has been focused on the building of simplified interpretable models as approximations of complex decision-making functions~\cite{Mittelstadt2019}. 
However, few works consider social science theories of explanation. For example, Miller suggests humans prefer contrastive explanations, or explanations that revolve around counterfactuals~\cite{miller2017explanation}. Specifically, humans tend to ask not why an event P happens, but why an event P happens instead of some event Q. Understanding this contrast of events is more important to the human user than statements of probabilities or lists of total causes. 

In this paper, we draw insights from the social sciences and formalize the notion of ``contrastive explanations'' in the context of robotic planning based on Markov decision processes (MDPs), which is a popular modeling formalism for representing abstract robotic mission plans~\cite{Thrun05}.
Our goal is to explain action choices in a planned robotic route, which can be computed as the optimal MDP policies using reinforcement learning~\cite{sutton2018reinforcement} or formal methods~\cite{lacerda2019probabilistic}.
More specifically, we focus on three key factors of contrastive explanations: 
\emph{selectiveness} (e.g., choosing the most relevant events)~\cite{miller2017explanation}, 
\emph{constrictiveness} (e.g., numbering how many future possible actions than an action causes)~\cite{GIROTTO1991111},
and \emph{responsibility} (e.g., rating how important an action is in causing an event) \cite{Halpern10.1093/lpr/mgu020}.
Different combinations of factors allow an explanation to control information specificity and support provided for events and actions.

\begin{figure}[ht]
     \centering
     \includegraphics[scale=0.12]{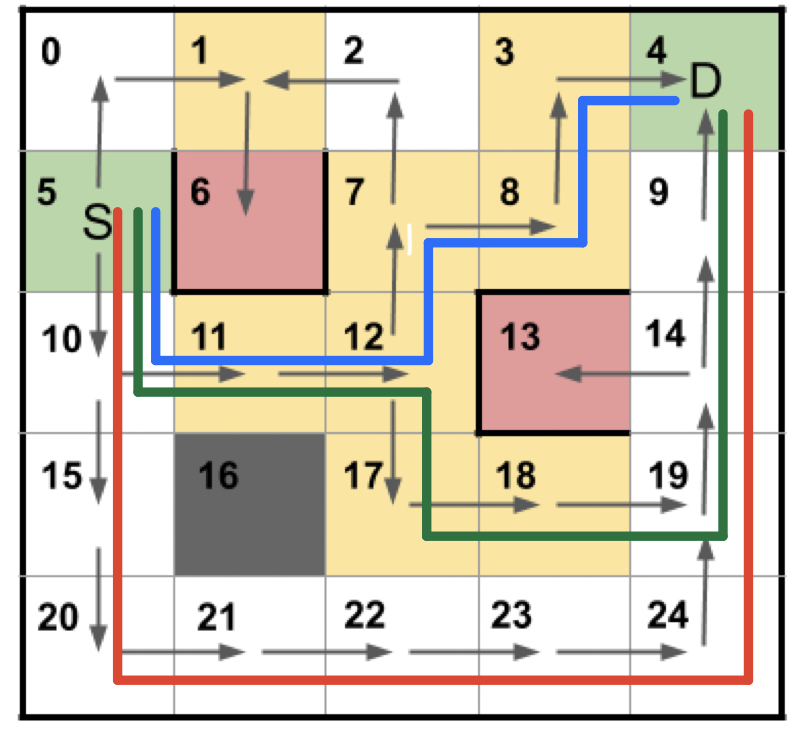}
     \caption{An example grid map for robotic planning. Green grids: start (S) and destination (D). Black grids: buildings. Red grids: dead-ends. Yellow grids: urban roads. White grids: highways. }
     \vspace{-10pt}
     \label{fig:grid}
 \end{figure}

\startpara{Motivating Example}
Consider the route planning for a robot navigating in a grid map as shown in Figure~\ref{fig:grid}.
There are three possible routes from the start (S) to the destination (D) highlighted in different colors.
The robot may take different routes, depending on the trade-offs of different objectives (e.g., minimizing the total route distance to destination, minimizing the risk of colliding with pedestrians or cyclists).
A naive way to explain a route is to generate a sentence for each action the robot takes at every state using a structured language template (e.g., ``We move east at grid 10.''), and then concatenate these sentences following the sequence of states in the route. 
However, it would be tedious if not infeasible to explain the robotic action in every state following the route, especially for large MDP models with hundreds of thousands of states. 
Therefore, we select a handful of critical states and only explain actions on those states. 
In addition to explain what action is taken in a state, we also explain why the action is taken by comparing it to alternative actions in terms of constrictiveness (e.g., ``We move east at grid 10 because it leads to the most flexible future route.'') and responsibility (e.g., ``We move east at grid 10 because it leads to the shortest route.'').

\startpara{Contributions}
We summarize the major contributions of this paper as follows:
\begin{enumerate}
    \item A formalization of contrastive explanations for MDPs based on three key factors (selectiveness, constrictiveness, and responsibility).
    \item A prototype implementation to automatically generate contrastive explanations for robotic planning based on MDPs.
    \item A user study with 100 participants on the Amazon Mechanical Turk platform to investigate the user understanding, trust and preference of contrastive explanations. 
\end{enumerate}

\startpara{Related Work}
When applied to AI-based systems, the finding of counterfactuals is often treated as a search or optimization problem \cite{Mittelstadt2019}. However, a counterfactual must be relevant to the system context or it will not produce an explanation that is understandable for the user. Additionally, counterfactuals can be isolated through the use of modeling by providing concise descriptions of system behavior\cite{kim_muise_shah_agarwal_shah_2019, miller2018contrastive, han_katoen_berteun_2009,1f738812b52b4a4c98b49d6369c23124}. Furthermore, explanation creation and policy transparency can be based in finding critical states, or the most important states, when reduction of the explanation is necessary \cite{huang2018establishing}.

The explanations provided by an AI-based decision-making system must deal with the significant trade-off between what the system is trying to accomplish and what the users need to understand the decisions made fully \cite{feng2016synthesis}. Balancing these trade-off increases system interpretability and user accessibility \cite{Lieaay6276}. So, when creating an explanation, all possible explanatory factors and support must be chosen carefully to maximize explaninee understanding and minimize explainee burden.

Explicitly, the generation of explanations for robotic planning through structured language templates has been done in work such as \cite{Hayes10.1145/2909824.3020233,feng2018counterexamples}. However, none of the previous works have produced contrastive explanations using selectiveness through the identification of critical states, responsibility, and constrictiveness, even though social science points to these factors as valid ways to increase explanation effectiveness.  

\section{Preliminaries}
In this section, we provide the necessary background on Markov decision processes (MDPs), 
which have been popularly used as a modeling formalism in robotic planning~\cite{Thrun05}.
Formally, an MDP model is a tuple $\M = (S, s_0, A, \delta, r)$, 
where $S$ is a finite set of \emph{states}, 
$s_0\in S$ is an \emph{initial state}, 
$A$ is a set of \emph{actions},
$\delta: S\times A\times S \to [0,1]$ is a \emph{transition relation} mapping each state-action pair to a probability distribution over $S$,
and $r: S\times A \times S \to \mathbb{R}$ is a \emph{reward function}.
At each MDP state $s$, first an action $a \in A$ is chosen nondeterministically based on an MDP \emph{policy} $\sigma: S \to A$, 
then a success state $s'$ is chosen with the probability $\delta(s,a,s')$.

Given an MDP model for robotic planning, there are many different methods for computing an optimal policy.
For example, various reinforcement learning techniques~\cite{sutton2018reinforcement} can compute an optimal MDP policy with the goal of maximizing the cumulative reward. 
In recent years, there are also increasing interests in applying formal methods to synthesize robotic plans subject to a rich set of MDP properties (e.g., probabilistic reachability, safety properties, liveness properties) expressed in temporal logic specifications~\cite{lacerda2019probabilistic}.  
Our approach is generally applicable for explaining any MDP policy, and is orthogonal to whether the policy is computed by reinforcement learning or formal methods. 

\begin{example}
We build an MDP model based on the grid map shown in Figure~\ref{fig:grid}.
The state space $S$ is defined by the grids. There are 25 states in total.
The initial state is grid 5 which is labeled with S. 
There are four actions in $A$: move north, move east, move south, and move west. 
We assume that, due to sensor uncertainty, the robot would perform an intended action correctly with probability 0.9 and get stuck in the same grid with probability 0.1.
An example transition relation is $\delta(g_5,\mathsf{south},g_{10})=0.9$ and $\delta(g_5,\mathsf{south},g_5)=0.1$.
We define a reward function $r$ for counting the total distance (e.g., number of grids) traveled for the robot to reach the destination. For example, $r(g_5,\mathsf{south},g_{10})=1$.

\end{example}

\section{Contrastive Explanations}

In this section, we formalize the three key factors of contrastive explanations:
selectiveness, constrictiveness and responsibility,
and present methods to compute them.

\subsection{Selectiveness}\label{sec: selective}


%

It would be tedious or even infeasible to explain a robot's action at every single state along the planned route, especially for large MDP models that may contain hundreds of thousands of states. 
Indeed, according to \cite{Mittelstadt2019}, explanations should be selective to reduce long causal chains to a cognitively manageable size for humans. 
To this end, we define the notion of \emph{critical states} in an MDP model and only explain actions in those states.
Intuitively, a critical state is where the choice of actions would greatly affect the MDP policies and their performance. 
For example, in grid 14 of Figure~\ref{fig:grid}, moving north is more likely to reach the destination while moving west would reach a dead end.  
Given a pair of state $s$ and action $a$ in an MDP model, we define the impact of this state-action pair as
$$\omega(s,a) = \sum_{s'\in S} \delta (s,a, s')\cdot \rho_{s'}$$
where $\delta (s,a, s')$ is the transition probability and 
$\rho_{s'}$ is the MDP property value (e.g., maximum cumulative reward) at a success state $s'$.
Then, we can obtain a pair of values for each state $s$
to measure the best/worst impact of different enabled actions $A(s)$:
$$\lambda^{max}_s = \max_{a\in A(s)} \omega(s,a)$$
$$\lambda^{min}_s = \min_{a\in A(s)} \omega(s,a)$$
Formally, we define the set of \textbf{critical states} of an MDP model with the state space $S$ as
$$S_c = \{s\in S \ |\ (\lambda_s^{max} - \lambda_s^{min}) > \alpha\}$$
where $\alpha$ is a user-defined threshold.

\begin{example}
Following the MDP model defined in Example 1 and considering a threshold $\alpha=0$ for the total distance of reaching the destination, we can compute the set of critical states as $\{g_5, g_7, g_{10}, g_{12}, g_{14}\}$.
We use grid 10 as an example to show the computation procedure. 
There are two enabled actions in grid 10: move east or move south.
Assume that $\rho_{g_{10}}=6.666, \rho_{g_{11}}=5.555, \rho_{g_{15}}=9.999$,
which represent the total expected distance of starting from grid 10, grid 11, and grid 15 to reach the destination, respectively. We have
$$\omega(g_{10}, \mathsf{east}) = 0.9\times5.555 + 0.1\times6.666 = 5.667$$
$$\omega(g_{10}, \mathsf{south}) = 0.9\times9.999 + 0.1\times6.666 = 9.667$$
$$\lambda_{g_{10}}^{max}-\lambda_{g_{10}}^{min} = 9.667 - 5.667 = 4 > 0$$
Thus, grid 10 is a critical state. 
\end{example}

\subsection{Constrictiveness}
In social sciences, a decision is said to be more ``constrictive'' if choosing it causes less possible future decisions \cite{miller2017explanation}. Overall, humans are more interested in less constrictive actions as we often have more future control when we choose them \cite{GIROTTO1991111}. Over time, actions tend to become more constrictive as a goal is reached.
In this paper, we interpret \emph{constrictiveness} as a measurement of how much an action would affect the flexibility in terms of the number of critical decision points left in the future route. 
Intuitively, more decision points lead to more flexibility for the robot to reroute, hence it is considered less constrictive and is preferred as time passes.
Given a state $s$ in an MDP model, we can construct a expectimax-like search tree~\cite{russell2010artificial} by taking the state $s$ as the root node, spanning with edges labeled with action $a$ leading to a set of children nodes $s'$ if the transition probability $\delta(s,a,s')>0$ until reaching target destination states.
We define the \textbf{constrictiveness} value of choosing an action $a$ in an MDP state $s$ as the number of critical decision points left in possible future routes by traversing the search tree $\T(s,a)$. Formally, 
$$\varepsilon(s,a) = \sum_{s\in \T(s,a) \cap S_c} A(s)$$


\begin{example}
\begin{figure}[t]
    \centering
    \subcaptionbox{$\T(g_{10}, \mathsf{south})$\label{fig:tree_10s}}{\includegraphics[scale=0.15]{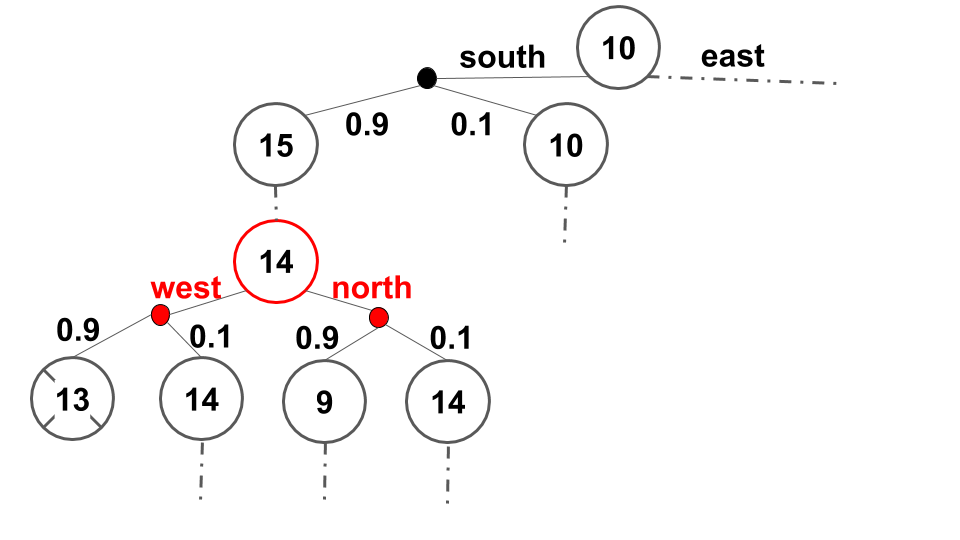}}\hspace{0em}%
    \subcaptionbox{$\T(g_{10}, \mathsf{east})$\label{fig:tree_10e}}{\includegraphics[scale=0.15]{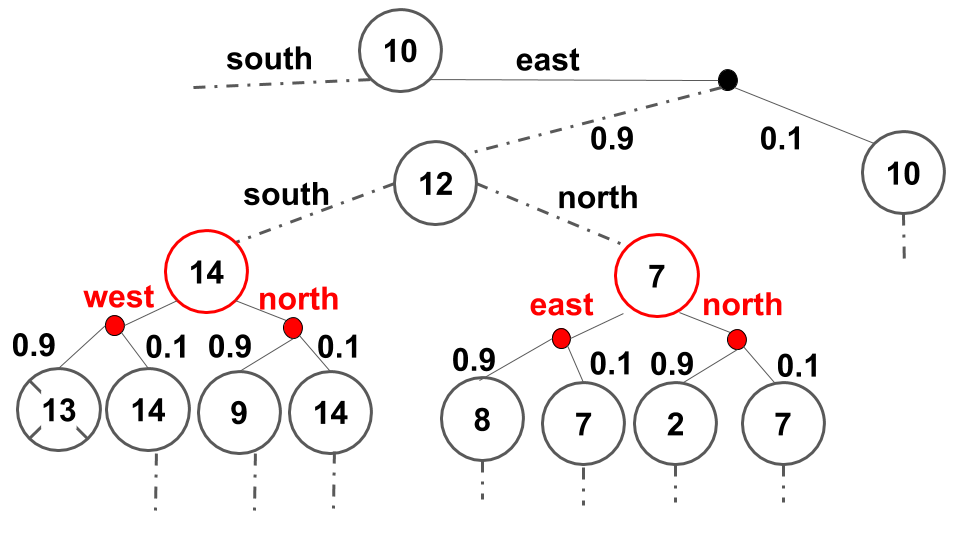}}
    \caption{Example search trees for grid 10}\label{fig:tree}
    \vspace{-5pt}
\end{figure}
Figure~\ref{fig:tree} shows two example search trees $\T(g_{10}, \mathsf{south})$ and $\T(g_{10}, \mathsf{east})$.
There is only one critical state $g_{14}$ in the tree $\T(g_{10}, \mathsf{south})$, with two enabled actions; thus
$\varepsilon (g_{10}, \mathsf{south}) = 2$.
And for the tree $\T(g_{10}, \mathsf{east})$, there are four future critical state-action pairs highlighted in red, that is $\varepsilon (g_{10}, \mathsf{east}) = 4$. 
This suggests moving east is more flexible with more critical decision points in future routes (i.e., less constrictive) than moving south, and thus is preferred. 
\end{example}

\subsection{Responsibility}
In social sciences, an action is said to be more ``responsible" if it changes the outcome more by removing that action from the current chosen path \cite{miller2017explanation}. 
Humans tend to be more interested in actions that hold a higher responsibility as it measures how much influence an action has over the final outcome \cite{Mittelstadt2019}.  
In this paper, we interpret responsibility as the measurement of an action's relative impact on the MDP property value compared with other actions enabled in the same state. 
Formally, we define the \textbf{responsibility} value of an action $a$ in an MDP state $s$ as
$$\zeta(s,a) = \omega(s,a) - \lambda_s^{min}$$
where $\omega(s,a)$ and $\lambda_s^{min}$ are defined in the Section~\ref{sec: selective}. 

\begin{example}
Following the previous Example 2, we know that 
$\omega(g_{10}, \mathsf{south})=9.667$, $\omega(g_{10}, \mathsf{east})=5.667$, and $\lambda_{g_{10}}^{min}=5.667$.
We can compute the responsibility value
$\zeta(g_{10}, \mathsf{south})=9.667-5.667=4$ and
$\zeta(g_{10}, \mathsf{east})=5.667-5.667=0$.
Thus, moving south is more responsible to the total distance, comparing with moving east.
Since we would prefer shorter route, moving east would be more preferable at grid 10. 
\end{example}

\section{User Study Design}\label{experiment}

\startpara{Experiment Domain} We designed a user study to evaluate the effectiveness of selectiveness, constrictiveness, and responsibility in contrastive explanations. For this study, we recruited 100 individuals with a categorical age distribution of 3 (0-17); 12 (18-24); 57 (25-34); 20 (35-49); 0 (50-64); and 2 (65+) using Amazon Mechanical Turk. We asked them to evaluate different types of explanations. Users were presented with 3 different 10-by-10 grid maps, each containing an optimal route from a start state to a finish state. Each route was presented with 7 different explanations about the robotic actions taken within it. An example of each of these explanations can be seen in Table \ref{tab:table-name}.

\begin{table*}[!t] \centering
\scalebox{0.86}{%
\begin{tabular}{|l|p{0.6\linewidth}|}
\hline
\textbf{Explanation Type} & \textbf{Explanation Example Text} \\ \hline
No Explanation & N/A \\ \hline
Naive Explanation (One State) & We move east at grid 10.       \\ \hline
Responsibility Explanation &    We move east at grid 10 because it leads to the shortest route.   \\ \hline
Constrictive Explanation & We move east at grid 10 because it leads to the most flexible future route.      \\ \hline
Naive Explanation (Entire Path) & First, we move south at grid 5. Next, we move east at grid 10. Then, we move east at grid 11. Next, we move north at grid 12. Then, we move east at grid 7. Next, we move north at grid 8. Finally, we move east at grid 3.      \\ \hline
Selective Explanation & First, we move south at critical grid 5. Then, we move east at critical grid 10. Next, we move north at critical grid 12. Finally, we move east at critical grid 7. All other decisions result in equivalent routes.\\ \hline
Contrastive Explanation (All Factors) & First, we move south at critical grid 5 because it leads to the shortest and most flexible future route. Then, we move east at critical grid 10 because it leads to the shortest and most flexible future route. Next, we move north at critical grid 12 because it leads to the shortest route. Finally, we move east at critical grid 7 because it leads to the shortest route. All other decisions result in equivalent routes.       \\ \hline
\end{tabular}}
\caption{\label{tab:table-name}Example of different explanations presented to users based on the grid map in Figure~\ref{fig:grid}}
\end{table*}

\startpara{Independent Variables} Each explanation was evaluated by the user on the level that the user understood the information presented by the explanation and the level that the user trusted that the information was correct. Both levels were measured using a 5-point Likert scale. Users were also asked to choose the explanation that they preferred out of several different groupings of explanations. Our independent variables included explanation type and explanation factors.
\startpara{Dependent Measures} The main subjective dependent variables were user understanding, user trust, and user preference. We measured time spent accessing the explanation as an objective dependent variable as well.

\startpara{Hypothesis} 
We have the following three hypotheses for this user study.

\textbf{H1.} We hypothesize that the use of selectiveness, responsibility, and constrictiveness in contrastive explanations will increase user understanding of information. 

\textbf{H2.} We hypothesize that the use of selectiveness, responsibility, and constrictivenss in contrastive explanations will increase user trust in explanation correctness.

\textbf{H3.} We hypothesize that users will prefer contrastive explanations using selectiveness, responsibility, and constrictiveness over other types of naive explanations.

\section{Results}\label{results}
In the following, we discuss the results of our user study regarding three hypotheses.

\subsection{Regarding H1 about user understanding}


\begin{figure}[h]
    \centering
    \includegraphics[scale=0.4]{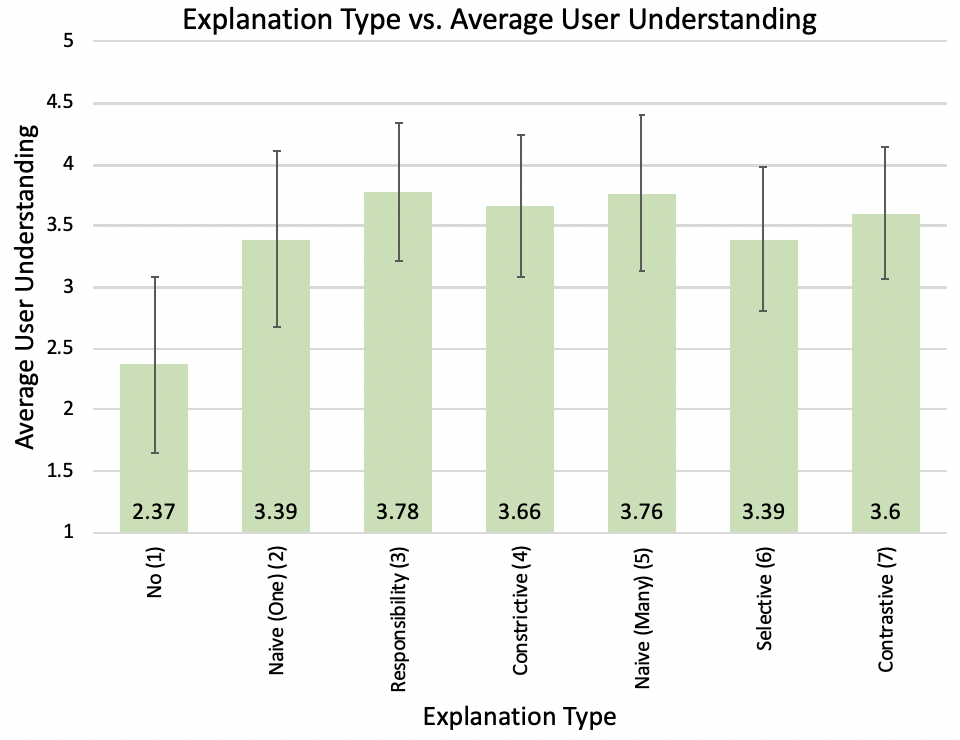}
    \caption{Display of Average User Understanding. 
}
    \label{fig:understanding}
\end{figure}

\begin{figure}[h]
    \centering
    \includegraphics[scale=0.4]{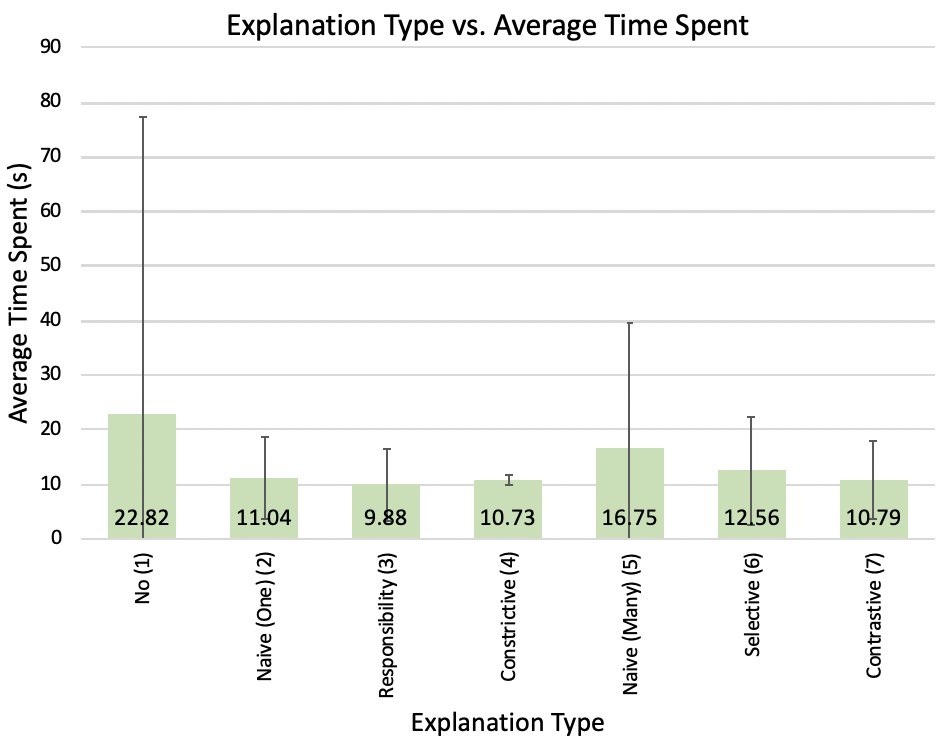}
    \caption{Display of Average User Time Spent}
    \label{fig:time}
\end{figure}

We begin by analyzing user understanding shown in Figure \ref{fig:understanding}. By a One-Way ANOVA ($\alpha$ = 0.05) test, F(6,2093) = 50.39, p $\leq$ 0.00001, the statistical differences between data shown is significant. As expected, presenting a user with no explanation allows for little understanding of the presented information. However, the introduction of responsibility or constrictive justification in the explanation increases user understanding of why actions are taken. Thus, users find it easier to understand an explanation if the justification for an action is presented alongside the action instead of just presenting the action. 

When dealing with selectiveness of an explanation, things are not as straight forward. This survey found that user understanding is decreased as the number of states explained was decreased to only the most critical states. Thus, a naive explanation is more effective in creating an overall understanding of the map than a selective one. However, we can also define user understanding in terms of cognitive burden, or the amount of time or energy the user must expend on processing the explanation. This factor is especially important in applications that are time-sensitive, such as autonomous vehicles, where the user has a short time to process the explanation and make a critical decision. Figure \ref{fig:time} shows the average time that users spent answering the survey questions regarding each explanation. An One-Way ANOVA test ($\alpha$ = 0.05), F(6,2093) = 2.9967, p = 0.0064, proves the statistical difference between the data for average time spent is significant. 
The use of a selective explanation decreases the amount of time that the user needs to understand the information given over its naive counterpart. The high standard deviation of the naive explanation shows that this is especially true for some users. So, selective explanations may impart less information, but they also decrease the amount of time needed to process that information. This may not be an important factor in a small example such as a 10-by-10 route map, but it is safe to project that as the number of states grows the importance of explanation selectiveness would increase as well, especially in models containing millions of possible states.

A contrastive explanation combining all three factors does not increase user understanding as we hypothesized in H1. This may be due in part to the selective factor that we discussed above. However, it did greatly decreased the amount of time needed for users to process the information over the naive explanation. Thus, contrastive explanation could be effective in increasing user understanding and decreasing cognitive burden in users.

\noindent\fbox{%
    \parbox{\linewidth}{%
        \textbf{Result 1:} The use of responsibility and constrictiveness increase understanding, while selectiveness decreases user understanding. Overall, contrastive explanations increase understanding and decrease cognitive burden.
    }%
}

\subsection{Regarding H2 about user trust}

\begin{figure}[t]
    \centering
    \includegraphics[scale=0.4]{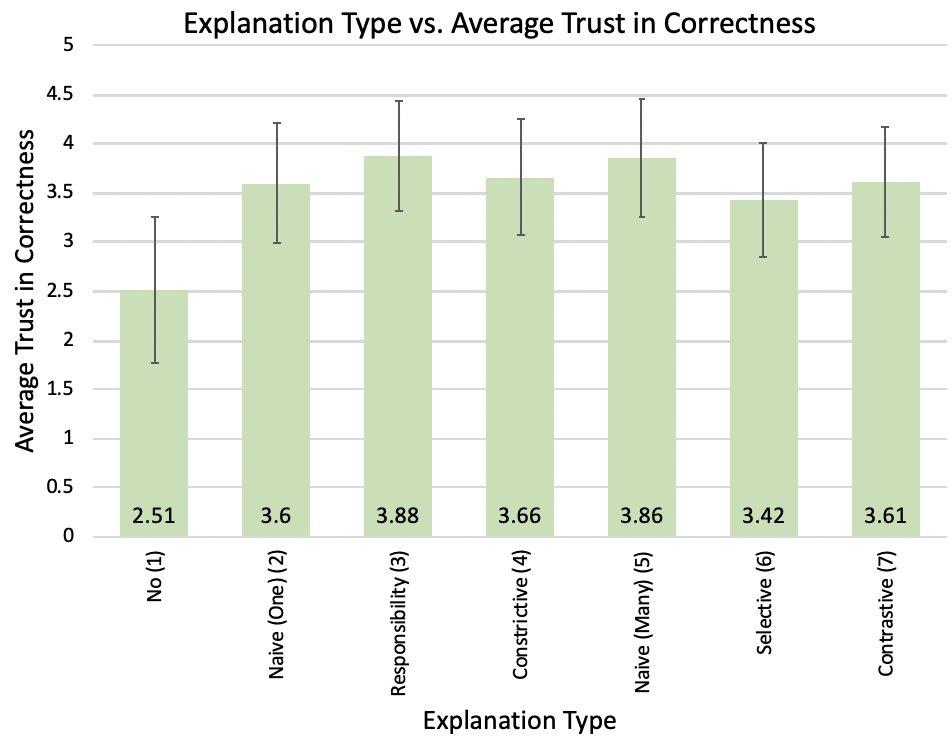}
    \caption{Display of Average User Trust.
}
    \label{fig:trust}
\end{figure}

Users not only need to understand the explanations presented, but they also need to trust that these explanation are correct. This can be achieve by providing relevant and necessary justification. Figure \ref{fig:trust} shows the average user trust in explanation correctness compared to the explanation type. By One-Way ANOVA ($\alpha$ = 0.05), F(6,2093) = 211.60, p $\leq$ 0.00001, the statistical difference between explanation trust averages is significant.

Providing no explanation to the user gives little trust to the system. However, using an explanation that provides support through action responsibility significantly increases user trust over a naive explanation. Yet, a constrictive explanation provides a significantly less increase in user trust than its responsibility counterpart, nearly offering the same amount of trust as the naive explanation. This may be due to the less direct connection of the constrictiveness justification compared to responsibility in reaching the destination state. It seems that selectiveness gives little help in increasing user trust as well. Even presented with the fact that only the states presented are important to reaching the established goal, users trust explanations that present more information about route actions. This may be because the naive explanation appears to have more information and thus more support even though this is not correct.\par
Additionally, putting all three factors together into a larger contrastive explanation seems to decrease the effectiveness of explanation in gaining user trust. This is most likely due to the integration of the selectiveness factor and the decrease in the number of states explained. However, the use of the responsibility and constrictive justification does help to establish more trust in a larger contrastive explanation bring the overall trust in the larger contrastive explanation up on average compared to a naive explanation.

\noindent\fbox{%
    \parbox{\linewidth}{%
        \textbf{Result 2:} The use of responsibility increases user trust, while the use of selectiveness decreases this factor. Constrictiveness has no effect. Overall, contrastive explanations increase user trust using responsibility justification.
    }%
}

\subsection{Regarding H3 about user preference}

Users not only need to understand and trust explanations, but they also must "like" them as they social entity subject to human preference. Someone preferring an explanation may make it more effective, as it may meet their needs of explanation justification and length better than other explanations. We found that users prefer responsibility explanations (46\%) and constrictive explanations (48\%) more often than their naive explanation counterparts (38\%, 32\% respectively). This is most likely due to the fact that people prefer explanations with some type of contrastive justification than just the presentation of actions. However, when dealing with selective explanations (26.3\%), users prefer a naive explanation (57.7\%) which presents each state and its action instead on one present the actions performed at critical states. Thus, users prefer to see more information, even if that information is not necessarily critical. Furthermore, we found that users prefer explanations using only responsibility (20.7\%), constrictive (27\%), or selective (23.7\%) factors almost just as often as contrastive explanations (28.7\%) combining all three factors. This in part may be due to user aversion to explanations utilizing selectiveness, but it also might point to large range of user preference that needs to be addressed in the creation of personalized explanations.

\noindent\fbox{%
    \parbox{\linewidth}{%
    {\textbf{Result 3:} Users prefer responsibility and constrictiveness explanations over naive explanations, but do not prefer selective explanations. Users prefer contrastive explanations at the same rate as single factor explanations.}
    }%
}

\section{Conclusion and Future Work}\label{conclusion}

In this paper, we present methods to compute contrastive explanations with three key factors (selectiveness, constrictiveness and responsibility) for robotic planning based on Markov decision processes, drawing on insights from the social sciences.
A user study with 100 participants on the Amazon Mechanical Turk platform shows that our generated contrastive explanations can improve user understanding and trust of autonomy, while reducing cognitive burden. 
In the future, we plan to further investigate methods of adapting explanations to an individual user's preferences and updating explanations in real-time based on user feedback.

\bibliographystyle{./bibliography/IEEEtran}
\bibliography{./bibliography/IEEEabrv,./bibliography/reference}

\begin{thebibliography}{10}
\providecommand{\url}[1]{#1}
\csname url@samestyle\endcsname
\providecommand{\newblock}{\relax}
\providecommand{\bibinfo}[2]{#2}
\providecommand{\BIBentrySTDinterwordspacing}{\spaceskip=0pt\relax}
\providecommand{\BIBentryALTinterwordstretchfactor}{4}
\providecommand{\BIBentryALTinterwordspacing}{\spaceskip=\fontdimen2\font plus
\BIBentryALTinterwordstretchfactor\fontdimen3\font minus
  \fontdimen4\font\relax}
\providecommand{\BIBforeignlanguage}[2]{{%
\expandafter\ifx\csname l@#1\endcsname\relax
\typeout{** WARNING: IEEEtran.bst: No hyphenation pattern has been}%
\typeout{** loaded for the language `#1'. Using the pattern for}%
\typeout{** the default language instead.}%
\else
\language=\csname l@#1\endcsname
\fi
#2}}
\providecommand{\BIBdecl}{\relax}
\BIBdecl

\bibitem{arrieta2020explainable}
A.~B. Arrieta, N.~D{\'\i}az-Rodr{\'\i}guez, J.~Del~Ser, A.~Bennetot, S.~Tabik,
  A.~Barbado, S.~Garc{\'\i}a, S.~Gil-L{\'o}pez, D.~Molina, R.~Benjamins
  \emph{et~al.}, ``Explainable artificial intelligence (xai): Concepts,
  taxonomies, opportunities and challenges toward responsible ai,''
  \emph{Information Fusion}, vol.~58, pp. 82--115, 2020.

\bibitem{Gunningeaay7120}
D.~Gunning, M.~Stefik, J.~Choi, T.~Miller, S.~Stumpf, and G.-Z. Yang,
  ``Xai{\textemdash}explainable artificial intelligence,'' \emph{Science
  Robotics}, vol.~4, no.~37, 2019.

\bibitem{adadi_berrada_2018}
A.~Adadi and M.~Berrada, ``Peeking inside the black-box: A survey on
  explainable artificial intelligence (xai),'' \emph{IEEE Access}, vol.~6, p.
  52138–52160, 2018.

\bibitem{goebel2018explainable}
R.~Goebel, A.~Chander, K.~Holzinger, F.~Lecue, Z.~Akata, S.~Stumpf,
  P.~Kieseberg, and A.~Holzinger, ``Explainable ai: the new 42?'' in
  \emph{International Cross-Domain Conference for Machine Learning and
  Knowledge Extraction}.\hskip 1em plus 0.5em minus 0.4em\relax Springer, 2018,
  pp. 295--303.

\bibitem{Mittelstadt2019}
B.~Mittelstadt, C.~Russell, and S.~Wachter, ``Explaining explanations in ai,''
  in \emph{Proceedings of the Conference on Fairness, Accountability, and
  Transparency}.\hskip 1em plus 0.5em minus 0.4em\relax ACM, 2019, pp.
  279--288.

\bibitem{miller2017explanation}
T.~Miller, ``Explanation in artificial intelligence: Insights from the social
  sciences,'' 2017.

\bibitem{Thrun05}
S.~Thrun, W.~Burgard, and D.~Fox, \emph{Probabilistic Robotics (Intelligent
  Robotics and Autonomous Agents)}.\hskip 1em plus 0.5em minus 0.4em\relax The
  MIT Press, 2005.

\bibitem{sutton2018reinforcement}
R.~S. Sutton and A.~G. Barto, \emph{Reinforcement learning: An introduction},
  2018.

\bibitem{lacerda2019probabilistic}
B.~Lacerda, F.~Faruq, D.~Parker, and N.~Hawes, ``Probabilistic planning with
  formal performance guarantees for mobile service robots,'' \emph{The
  International Journal of Robotics Research}, vol.~38, no.~9, pp. 1098--1123,
  2019.

\bibitem{GIROTTO1991111}
V.~Girotto, P.~Legrenzi, and A.~Rizzo, ``Event controllability in
  counterfactual thinking,'' \emph{Acta Psychologica}, vol.~78, no.~1, pp. 111
  -- 133, 1991.

\bibitem{Halpern10.1093/lpr/mgu020}
J.~Y. Halpern, ``{Cause, responsibility and blame: a structural-model
  approach},'' \emph{Law, Probability and Risk}, vol.~14, no.~2, pp. 91--118,
  01 2015.

\bibitem{kim_muise_shah_agarwal_shah_2019}
J.~Kim, C.~Muise, A.~Shah, S.~Agarwal, and J.~Shah, ``Bayesian inference of
  linear temporal logic specifications for contrastive explanations,''
  \emph{Proceedings of the Twenty-Eighth International Joint Conference on
  Artificial Intelligence}, 2019.

\bibitem{miller2018contrastive}
T.~Miller, ``Contrastive explanation: A structural-model approach,'' 2018.

\bibitem{han_katoen_berteun_2009}
T.~Han, J.-P. Katoen, and D.~Berteun, ``Counterexample generation in
  probabilistic model checking,'' \emph{IEEE Transactions on Software
  Engineering}, vol.~35, no.~2, p. 241–257, 2009.

\bibitem{1f738812b52b4a4c98b49d6369c23124}
B.~Krarup, M.~Cashmore, D.~Magazzeni, and T.~Miller,
  ``\BIBforeignlanguage{English}{Model-based contrastive explanations for
  explainable planning},'' in \emph{\BIBforeignlanguage{English}{ICAPS 2019
  Workshop on Explainable AI Planning (XAIP)}}, 7.

\bibitem{huang2018establishing}
S.~H. Huang, K.~Bhatia, P.~Abbeel, and A.~D. Dragan, ``Establishing appropriate
  trust via critical states,'' in \emph{2018 IEEE/RSJ International Conference
  on Intelligent Robots and Systems (IROS)}.\hskip 1em plus 0.5em minus
  0.4em\relax IEEE, 2018, pp. 3929--3936.

\bibitem{feng2016synthesis}
L.~Feng, C.~Wiltsche, L.~Humphrey, and U.~Topcu, ``Synthesis of
  human-in-the-loop control protocols for autonomous systems,'' in \emph{IEEE
  Transactions on Automation Science and Engineering}.\hskip 1em plus 0.5em
  minus 0.4em\relax IEEE, 2016, pp. 450--462.

\bibitem{Lieaay6276}
X.~Li, Z.~Serlin, G.~Yang, and C.~Belta, ``A formal methods approach to
  interpretable reinforcement learning for robotic planning,'' \emph{Science
  Robotics}, vol.~4, no.~37, 2019.

\bibitem{Hayes10.1145/2909824.3020233}
B.~Hayes and J.~A. Shah, ``Improving robot controller transparency through
  autonomous policy explanation,'' in \emph{Proceedings of the 2017 ACM/IEEE
  International Conference on Human-Robot Interaction}, ser. HRI ’17, p.
  303–312.

\bibitem{feng2018counterexamples}
L.~Feng, M.~Ghasemi, K.-W. Chang, and U.~Topcu, ``Counterexamples for robotic
  planning explained in structured language,'' in \emph{2018 IEEE International
  Conference on Robotics and Automation (ICRA)}.\hskip 1em plus 0.5em minus
  0.4em\relax IEEE, 2018, pp. 7292--7297.

\bibitem{russell2010artificial}
S.~Russell, S.~Russell, P.~Norvig, and E.~Davis, \emph{Artificial Intelligence:
  A Modern Approach}.\hskip 1em plus 0.5em minus 0.4em\relax Prentice Hall,
  2010.

\end{thebibliography}

\end{document}